\documentclass[letterpaper, 10 pt, conference]{ieeeconf}  % Comment this line out if you need a4paper

\IEEEoverridecommandlockouts                              % This command is only needed if 
\usepackage[T1]{fontenc}
\usepackage{subfig}
\usepackage{color}
\usepackage{hyperref}
\usepackage{breakurl}
\usepackage{pifont}% http://ctan.org/pkg/pifont

\title{\LARGE \bf
Alleviating the Burden of Labeling: \\Sentence Generation by Attention Branch Encoder--Decoder Network
}

\author{Tadashi Ogura$^{1}$, Aly Magassouba$^{1}$, Komei Sugiura$^{1,2}$, Tsubasa Hirakawa$^{3}$, 
Takayoshi Yamashita$^{3}$, \\Hironobu Fujiyoshi$^{3}$, and Hisashi Kawai$^{1}$% <-this % stops a space
\thanks{$^{1}$Authors are with the National Institute of Information and Communications
Technology,
3-5 Hikaridai, Seika, Soraku, Kyoto 619-0289, Japan.
        {\tt\small firstname.lastname@nict.go.jp}}%
\thanks{$^{2}$Author is with Keio University, 3-14-1 Hiyoshi, Kohoku, Yokohama, Kanagawa 223-8522, Japan.
         {\tt\small firstname.lastname@keio.jp}
}
\thanks{$^{3}$Authors are with Chubu University, 1200 Matsumotocho, Kasugai, Aichi 487-8501, Japan.
         {\tt\small \{hirakawa@mprg.cs, takayoshi@isc, fujiyoshi@isc\}.chubu.ac.jp}}
}

\usepackage{mymacros}
\usepackage{times} % assumes new font selection scheme installed
\usepackage{graphicx}

\let\labelindent\relax
\usepackage{enumitem}

\begin{document}

\maketitle
\thispagestyle{empty}
\pagestyle{empty}

%%%%%%%%%%%%%%%%%%%%%%%%%%%%%%%%%%%%%%%%%%%%%%%%%%%%%%%%%%%%%%%%%%%%%%%%%%%%%%%%
%\begin{abstract}
%Placing objects is a fundamental task for domestic service robots (DSRs). Most conventional methods detect free areas but do not predict the physical placeability (likelihood of success). Predicting placeability is challenging because it depends on the physical properties of the robot hardware, destination, obstacles, and the target object. Addressing this, we developed a CNN network with an attention mechanism to predict placeability on different concrete areas of the image, based on RGB-D images. We extend the method with GAN based data augmentation. Experimental results show that our approach significantly improved accuracy compared with baseline methods.
%\end{abstract}

% \begin{itemize}
%  % \item related  25
%  \item youtube
%  % \item video
%  % \item algorithm?
% \end{itemize}

\begin{abstract}
Domestic service robots (DSRs) are a promising solution to the shortage of home care workers.
However, one of the main limitations of DSRs is their inability to interact naturally through language.
Recently, data-driven approaches have been shown to be effective for tackling this limitation; however, they often require large-scale datasets, which is costly.
Based on this background, we aim to perform automatic sentence generation of fetching instructions: for example, ``Bring me a green tea bottle on the table.''
This is particularly challenging because appropriate expressions depend on the target object, as well as its surroundings.
In this paper, we propose the attention branch encoder--decoder network (ABEN), to generate sentences from visual inputs.
Unlike other approaches, the ABEN has multimodal attention branches that use subword-level attention and generate sentences based on subword embeddings.
In experiments, we compared the ABEN with a baseline method using four standard metrics in image captioning.
Results show that the ABEN outperformed the baseline in terms of these metrics.
\end{abstract}

% % 8/8

\vspace{-1mm}
\section{Introduction
\label{intro}
}
\vspace{-1mm}

% crucial, critical = very important

 The growth in the aged population has steadily increased the need for daily care and support. 
Domestic service robots (DSRs) that can physically assist people with disabilities are a promising solution to the shortage of home care workers~\cite{piyathilaka2015human,smarr2014domestic,iocchi2015robocup}.
This has boosted the need for standardized DSRs that can provide necessary support functions.

Nonetheless, one of the main limitations of DSRs is their inability to interact naturally through language. 
Indeed, most DSRs do not allow users to instruct them with diverse expressions.
Recent studies have shown that data-driven approaches are effective for handling ambiguous instructions \cite{anderson2018vision,wang2019reinforced,magassouba2018multimodal,magassouba2019understanding}.

Unfortunately, these approaches often require large-scale datasets, and are time-consuming and costly.
The main reason is the time that is required for human experts to provide sentences for images.
Hence, methods to augment or generate instructions automatically
% \Update to build large-scale datasets \Done 
could drastically reduce this cost and alleviate the burden of labeling from human experts.

\begin{figure}[t]
    \centering
    \includegraphics[width=\linewidth]{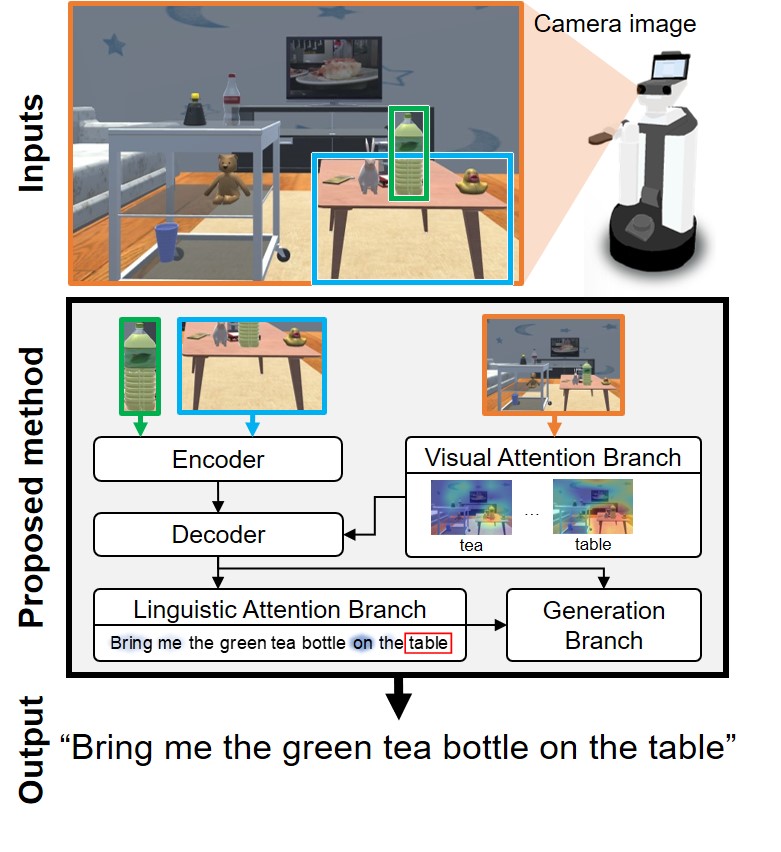}
    \vspace{-12mm}
    \caption{\small Overview of the ABEN: the ABEN generates fetching instructions from given input images.}
    \label{fig:eye_catch}
    \vspace{-6mm}
\end{figure}

Based on this background, we aim to perform automatic sentence generation of ``fetching instructions'' (instructions to the DSR to fetch items). 
This task involves generating a natural fetching instruction, given a target object in an image: for example, ``{\it Bring me a green tea bottle on the table.}'' Such an instruction often includes a referring expression, such as ``a green tea bottle on the table.''
A referring expression is an expression in which an object is described with regard to a landmark, such as ``table''.
This is particularly challenging because of the many-to-many mapping between language and the environment.

In this paper, we propose the attention branch encoder--decoder network (ABEN), which generates fetching instructions from visual inputs.
Fig.~\ref{fig:eye_catch} shows a schematic diagram of the approach.
The ABEN comprises a visual attention branch (VAB) and a linguistic attention branch (LAB), to attend both visual and linguistic inputs. 
An additional generation branch is introduced to generate sentences.
%A typical use case is to build a large-scale multimodal dataset. 
%Additionally, the method could be used in DSRs to produce visual explanations of actions.

The ABEN extends the attention branch network (ABN)~\cite{fukui2019attention} by introducing multimodal attention branches.
%The ABN is an image classifier, inspired by the class activation mapping (CAM) \cite{zhou2016learning} structure, that generates attention maps. The ABN include this structure as an attention branch that highlights the most salient portions in the image given a label to predict. 
In the ABN, attention maps are output by an attention branch, which highlights the most salient portions in the image, given a label to predict. 
In the ABEN, the attention map given by the VAB can serve as a visual explanation of the model, which is usually a black box.
Similarly, the attention map provided by the LAB can serve as an explanation of subword relationships.

The ABEN was inspired by our previous approach, the multimodal ABN (Multi-ABN) \cite{magassouba2019multimodal}, and shares its basic structure.
The main differences between the ABEN and the Multi-ABN include the subword-level attention used in the LAB and BERT\cite{devlin2018bert}-based subword embeddings used for sentence generation. A demonstration video is available at this URL\footnote{\protect\href{https://youtu.be/H7vsGmJaE6A}{https://youtu.be/H7vsGmJaE6A}}.

The main contributions of this paper are as follows:
\begin{itemize}
 \item The ABEN extends the Multi-ABN by introducing a linguistic branch and a generation branch, to model the relationship between subwords. 
 \item The ABEN combines attention branches and BERT-based subword embedding, for sentence generation.
\end{itemize}

\vspace{-2mm}
\section{Related Work
\label{related}
}
\vspace{-0.5mm}
There have been many attempts to construct communicative robots for manipulation tasks~\cite{iocchi2015robocup}. Recently, in some studies \cite{sugiura2010active,hatori2018interactively,shridhar2018interactive,magassouba2019understanding,magassouba2018multimodal}, linguistic inputs were processed along with visual information to handle the many-to-many mapping between language and the environment.

These studies have often used data-driven approaches that were originally proposed in the natural language processing (NLP) and computer vision communities. 
For instance, \cite{hatori2018interactively} proposed a method for predicting target objects from natural language in a pick-and-place task environment, using a visual semantic embedding model. Similarly, \cite{shridhar2018interactive} tackled the same type of problem using a two-stage model to predict the likely target from the language expression and the pairwise relationships between different target candidates. More recently, in a context related to DSRs, \cite{magassouba2019understanding} proposed the use of both the target and source candidates to predict the likely target in a supervised manner. In \cite{magassouba2018multimodal}, the placing task was addressed through a generative adversarial network (GAN) classifier that predicted the most likely destination from the initial instruction.

Nonetheless, these methods required large-scale datasets, which are seldom available in a DSR context because they require substantial labeling effort from human experts. Datasets such as RefCOCO~\cite{kazemzadeh2014referitgame} or MSCOCO~\cite{lin2014microsoft} are widely used in visual captioning, however they are not specifically designed for robots.  The Room-to-Room dataset~\cite{anderson2018vision} is a dataset designed for multimodal language understanding for navigation, however manipulation is not handled.
Conversely, a pick-and-place dataset such as PFN-PIC \cite{hatori2018interactively}, contains top-view images only, and does not handle furniture, and is therefore not suitable for DSRs.

To address this problem, a promising solution involves generating synthetic instructions  to label unseen visual inputs to augment such datasets. Moreover, such a method enables real-time task generation in simulators, where DSRs are instructed to fetch everyday objects in randomly generated environments.

Most studies use rule-based approaches to generate sentences (e.g. \cite{kunze2017spatial}),
% in which several algorithms are evaluated.
% 

% By using positional information in the image, some reference representations can be described by a rule-based approach.
% 
however, they cannot fully capture and reproduce the many-to-many mapping between language and the physical world. 
Indeed, handling natural sentences that include referring expressions is particularly challenging. 
Conversely, an end-to-end approach was used in \cite{dogan2019learning} for estimating spatial relations to describe an object in a sentence. 
Nonetheless, the set of spatial relations was limited to six and was hand-crafted. 
Unlike these studies, we target an end-to-end approach that does not require hand-crafted or rule-based methods. 
In our previous work \cite{magassouba2019multimodal}, we introduced the Multi-ABN, which generates fetching instructions by using a multimodal attention branch mechanism \cite{fukui2019attention}. The Multi-ABN is a long short-term memory (LSTM) that is enhanced by visual and linguistic attention branches.

This study extends the Multi-ABN by introducing subword-level attention, which has the benefit of interpretability, unlike the linguistic attention in \cite{magassouba2019multimodal}.

Our approach can model the relationship between the generated subwords.

Furthermore, unlike most sentence generation methods, our approach generates sentences via a BERT-based subword embedding~\cite{devlin2018bert} model, which was shown by \cite{magassouba2019understanding} to perform better than a word embedding model.

\begin{figure}[t]
    \centering
    \includegraphics[clip,width=\linewidth]{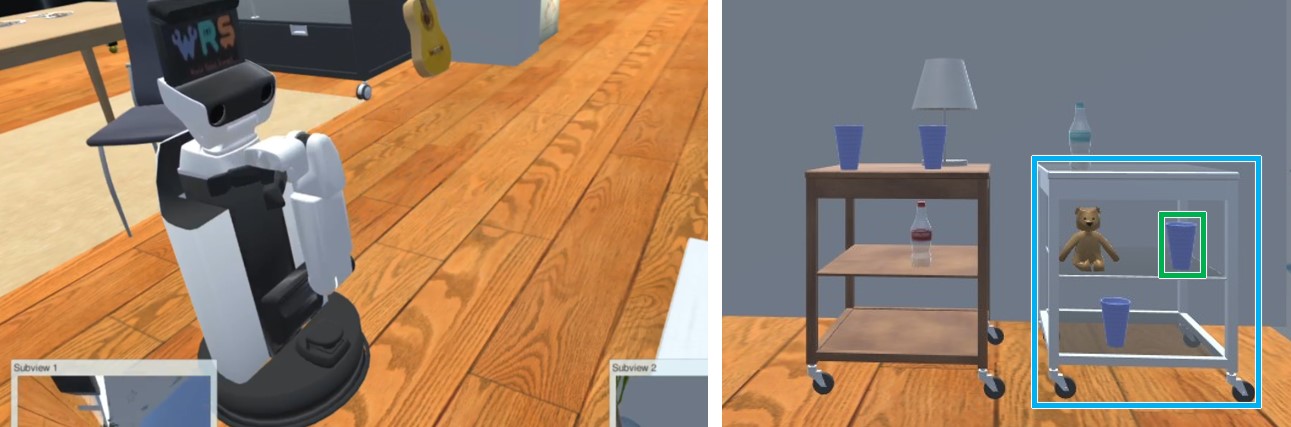}
    \vspace{-6mm}
    \caption{\small Left: Typical scene in which the DSR is observing everyday objects. Right: The camera image recorded from the DSR's position shown in the left-hand panel. The blue and green boxes represent the target (blue glass) and its source (metal wagon).  Typical instructions include ``Bring me the blue glass next to the teddy bear'' and  ``Bring me the blue glass on the same level as the teddy bear on the metal wagon.''}
    \label{fig:scene}
    \vspace{-6mm}
\end{figure}

Therefore, the architecture of  the ABEN extends the ABN with multimodal attention. Multimodal attention has been widely investigated in image captioning. Recent studies in multimodal language understanding have shown that both linguistic and visual attention are beneficial for question-answering tasks~\cite{nguyen2018improved,lei2019tvqa} or visual grounding~\cite{akbari2019multi,yu2018mattnet}.
Similarly, \cite{hori2017early} introduces an attention method that performs a weighted average of linguistic and image inputs.
In contrast to these attention mechanisms, attention branches are based on class activation mapping (CAM) networks~\cite{zhou2016learning}.
CAM focuses on the generation of masks that, overlaid onto an image, highlight the most salient area given a label. In the ABN, such a structure is introduced through an attention branch that generates attention maps to improve the prediction accuracy of the base network.
In the ABEN, visual and linguistic attention maps are generated to mask the visual input and the sequence of generated subwords.

\begin{figure*}
    \centering
    \includegraphics[clip,width=150mm]{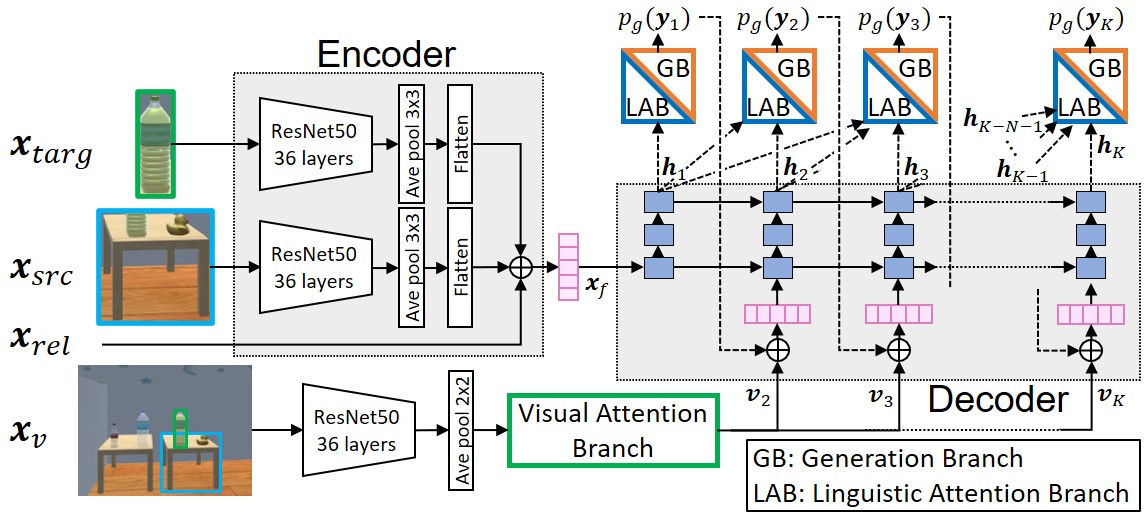}
    \vspace{-4mm}
    \caption{\small Structure of the ABEN. The ABEN comprises an encoder, a decoder, a visual attention branch, a linguistic attention branch, and a generation branch.}
    \label{fig:encoder-decoder}
    \vspace{-7mm}
\end{figure*}

Subwords have been widely used for machine translation \cite{schuster2012japanese,sennrich2016neural,kudo2018subword} as well as being used in most recent language models such as BERT, ALBERT~\cite{lan2019albert} or XLNet~\cite{yang2019xlnet}. These methods achieve state-of-the-art performance in many natural language understanding tasks. 
Combining such a method with the attention branch architecture to enrich sentence generation, is one of the main novelties of the ABEN. Indeed, traditionally in robotics, very simple embedding and language models have been used. For example, recent studies used simple skip-gram (e.g. \cite{hatori2018interactively,matthews2019word2vec,cohen2019grounding,shridhar2018interactive}). 

\section{Problem Statement
\label{sec:problem}
}

% Q.1
% The fetching task is one of the basic skills required for DSR.

%  I think that some important points are missing. You have to convince the reviewer that this task is challenging.
% The story should be like:
% 1. We want to generate fetching instructions for DSRs
% 2. Despite the many-to-many mapping of language and physical world, we want to generate sentence that are NOT AMBIGUOUS 
% 3. That means that an instruction should refer to only ONE object in the scene
% 4. Fig.1 should show a difficult case: For instance a scene where there is several time the same object -> How to generate an instruction for this case?
% 5. Hence, to disambiguate this configuration illustrated in Fig.1  we need to generate REFERRING EXPRESSIONS.
% 6. referring expressions are ...
% 7. To address this problem we use only visual inputs etc.. and no hand crafted-rules which makes it very challenging
% 8. The inputs are ...
% 9. We assume 2D scene. SO we do not handle behind/in front because etc...
% 9. To evaluate our approach we use standard...

\vspace{-0.5mm}
\subsection{Task description}
\vspace{-1mm}
% \begin{wrapfigure}{r}{40mm}
%     % \begin{center}
%         \includegraphics[clip,width=40mm]{fig/eye_catch/same_objects_v1.jpg}
%         \caption{A scene image where there is several time the same object.}
%         \label{fig:same_objects}
%     % \end{center}
% \end{wrapfigure}
% \begin{figure}
%     \centering
%     \includegraphics[clip,width=50mm]{fig/eye_catch/same_objects_v1.jpg}
%     % \includegraphics[bb=0 0 673 369,width=\linewidth]{fig/network_structure/lstm.jpg}
%     \caption{A scene image where there is several time the same object.}
%     \label{fig:same_objects}
% \end{figure}

The study targets natural language generation for DSRs.
Hereinafter, we call this task the fetching instruction generation (FIG) task.
A typical scenario is shown in Fig.~\ref{fig:scene}, in which the target to fetch is described in the instruction ``Bring me the blue glass on the same level as the teddy bear on the metal wagon.'' This example emphasizes the challenges of the FIG task, because each instruction should describe the targeted object uniquely. 

To avoid ambiguity, it is necessary to generate sentences including referring expressions, because there may be many objects of the same type. 
Referring expressions allow the targeted object to be characterized uniquely with respect to its surrounding environment. In Fig.~\ref{fig:scene}, the referring expression ``on the same level as the teddy bear on the metal wagon'' is needed to disambiguate the targeted object from others. 

This is particularly challenging because appropriate expressions depend on the target itself, as well as its surroundings. For instance, the target object in Fig.~\ref{fig:scene} can be described as ``glass near the bear doll'' and ``blue tumbler glass on the second level of the wagon'', in addition to many other candidate expressions. Therefore, it is necessary to handle the many-to-many mapping between language and the physical world.

The FIG task is characterized by the following:
\begin{itemize}
    \item \textbf{Input:} RGB image of an observed scene.
    \item \textbf{Output:} the most likely generated sentence for a given target and source.
\end{itemize}
The inputs of the ABEN are explained in detail in Section~\ref{method}.

% Q.5
We define the terms used in this paper as follows:
\begin{itemize}
    \item \textbf{Target:} an everyday object, e.g., bottles or fruit, that is to be fetched by the robot.
    \item \textbf{Source:} the origin of the target, e.g., furniture, such as shelves or drawers.
\end{itemize}

% We assume 2D scene. SO we do not handle behind/in front because etc...
In the FIG task, we assume that the two-dimensional bounding boxes of the target and the source are defined in advance. 
Furthermore, referring expressions related to depth perception (e.g., ``behind'' or ``in front'') are not addressed because no three-dimensional information is available.
%We assume that the position and size of the target and source are given beforehand. Such an assumption is reasonable because many object recognition and detection methods are available in the literature(e.g. \cite{redmon2016you}). Therefore, object recognition and detection is out of the scope of this study. 

% Q.9
% We use standard the natural language processing (NLP) metrics for sentence generation evaluation (BLUE, ROUGE, CIDEr, METEOR).
The evaluation of the generated sentences is based on several standard metrics---BLEU~\cite{papineni2002bleu}, ROUGE~\cite{lin-2004-rouge}, METEOR~\cite{banerjee2005meteor}, and CIDEr~\cite{vedantam2015cider}---that are commonly used for image captioning studies.  
Although they are imperfect, by using several metrics, we may overcome their limitation and assess better the quality of the sentences. In \cite{banerjee2005meteor}, BLEU and METEOR were reported to have a correlation of 0.817 and 0.964, respectively, with human evaluation. Furthermore, these metrics also allow us to compare our approach to existing methods. %thanks!

A simulated environment (see Fig. \ref{fig:scene}) is used to collect the image inputs. Indeed, because we aim to generate sentences in a wide range of configurations, using a simulated environment is effective for addressing these situations at a low cost.
Moreover, using a simulation has the advantage that the experimental results can be reproduced. 

As the simulated robot platform, we use a standardized DSR, namely  Human Support Robot (HSR) ~\cite{yamamoto2019development}.
Our simulator is based on SIGVerse~\cite{inamura2013development,mizuchi2017cloud}, which is an official simulator for HSR that provides a three-dimensional environment based on the Unity engine.

In the data collection phase, HSR~\cite{yamamoto2019development} navigates in procedurally generated environments with everyday objects. Thereafter, RGB images of target and source candidates are recorded using the camera, with which HSR is equipped.

\vspace{-1mm}
\section{Proposed Method
\label{method}
}
\vspace{-1mm}
\subsection{Novelty}
\vspace{-1mm}

To generate fetching instructions, the ABEN extends the Multi-ABN~\cite{magassouba2019multimodal} by introducing a subword generation architecture using BERT \cite{devlin2018bert} embedding in addition to subword-level attention. For this purpose, as shown in Fig.~\ref{fig:encoder-decoder}, the ABEN comprises an encoder (base network), a decoder, a VAB, and a LAB. 
The following characteristics of the ABEN should be emphasized:
\begin{itemize}
    \item Unlike the ABN~\cite{fukui2019attention}, which comprises a base network coupled with attention and perception branches, the ABEN follows an encoder--decoder structure (i.e., there is no perception branch) based on an LSTM network. 
    \item Unlike the Multi-ABN, fetching instructions are generated from a sequence of subwords with BERT encoding. %From this approach, the quality of the generated sentences is drastically improved as the mapping of language to the image, it is better captured by a richer vocabulary. This will be explained in the next sub section.
    \item The ABEN introduces the novel structures of linguistic attention branches and generation branches to allow a subword-level attention mechanism. Hence, the ABEN attention is fully interpretable, unlike that of the Multi-ABN which uses latent-space linguistic attention. 
    
\end{itemize}
%The proposed method generates a sequence $Y = \{ {\bf y}_1, {\bf y}_2 \dots {\bf y}_J \}$  where $J$ is the length of a generated sentence and ${\bf y}_j$ is an predicted word in the term of $j$.

\begin{table}[t]
\small
\caption{\small Difference between (a) typical word-tokens with pre-processing for rare and/or misspelled words and (b) sub-word tokenization.}
\vspace{-3mm}
\label{tab:word}
\centering
\begin{tabular}{l|c|c}
\hline
Expression &(a) &(b)\\
\hline
\hline
 topright object &topright, object& top, right, object \\
\hline  
sprayer &  $<$UNK$>$ & spray, er\\
\hline
grayis bottle & $<$UNK$>$ , bottle & gray, is, bottle \\
\hline
\end{tabular}
\vspace{-3mm}
\end{table}

\vspace{-2mm}
\subsection{Input and Subword Tokenization}
\vspace{-1mm}

Fig.~\ref{fig:encoder-decoder} shows the network structure of the ABEN. For a scene $i$, let us define our set of inputs $\textbf{x}_i$ as:
%Let us define our dataset $\textbf{X}$ as   $\textbf{X}=\{\textbf{x}_i|i=1, \cdots ,N \}$, where $\textbf{x}_i$ is the $i$-th input sample among a set of $N$ samples.
% The $i$-th sample $\textbf{x}_i$ in the dataset is defined as $\textbf{X}~=~\{\textbf{x}_i|i=1, \cdots ,N \}$.
\vspace{-0.5mm}
\begin{equation}
    \textbf{x}_i = \{\textbf{x}_v(i), \textbf{x}_{src}(i), \textbf{x}_{targ}(i), \textbf{x}_{rel}(i)\}.
\vspace{-0.5mm}
\end{equation}
For readability, we omit the index $i$ so that $\textbf{x}_i$ is simply written as $\textbf{x}$. Given this, the inputs are defined as follows:
\begin{itemize}
    \item $\textbf{x}_v$: the input scene as an RGB image.
    \item $\textbf{x}_{targ}$: the cropped image of the target in $\textbf{x}_v$
    \item $\textbf{x}_{src}$: the cropped image of the source in $\textbf{x}_v$
    \item $\textbf{x}_{rel}$: the relational features between $\textbf{x}_v$, $\textbf{x}_{targ}$, and $\textbf{x}_{src}$.
\end{itemize}
$\textbf{x}_{rel}$ comprises the position and size features of (a) the target relative to the source, (b) the target relative to the full image, and (c) the source relative to the full image.
Each of these relations is characterized by:
\vspace{-0.5mm}
\begin{equation}
    \textbf{r}_{l/m}=\left[\frac{x_l}{W_m}, \frac{y_l}{H_m}, \frac{w_l}{W_m}, \frac{h_l}{H_m}, \frac{w_l h_l}{W_m H_m}\right],
    \label{eq:x_rel}
\vspace{-0.5mm}
\end{equation}
where $x_l$, $y_l$, $w_l$, and $h_l$ denote the horizontal and vertical positions and the width and height, respectively, of the component $l$. $W_m$ and $H_m$ denote the width and height, respectively, of the component $m$. Consequently, the relation features are defined as $\textbf{x}_{rel} = \{\textbf{r}_{targ/src}, \textbf{r}_{targ/v}, \textbf{r}_{src/v}\}$ with dimension 15.
% $\textbf{x}_{src}$ are standardized in order to exhibit a zero mean and unit variance for each of them. The other image features are downscaled to fit in $(224 \times 224 \times 3)$ before being processed by a CNN to extract visual features.

In contrast to most methods for sentence generation, BERT-based subword embeddings, instead of classic word-based embedding, are used as the ground truth. 
BERT was pretrained on 3.5 billion words and is therefore robust against data sparseness regarding rare words. 
% The embedding model is then fine-tuned on the data set as the XXXX is trained.
In our previous work on multimodal language understanding, we introduced BERT-based subword embedding; this was one of the earliest applications of BERT in robotics~\cite{magassouba2019understanding}. 
It has been reported that BERT-based subword embedding functioned better than simple word-based embedding for the PFN-PIC dataset~\cite{hatori2018interactively}.
In many NLP studies, BERT and other Transformer-based approaches have been applied successfully to challenging tasks. 
For domain adaptation, we can additionally fine-tune a BERT-based model that is pre-trained on a large-scale dataset.

Furthermore, subword tokenization \cite{wu2016google} is robust against the misspelling words. Indeed, a matching is still possible in subword units. As illustrated in Table \ref{tab:word}, a the word `grayish' misspelled as `grayis' can still be matched with the subword 'gray' which is impossible with classic word embedding. As a result, the subword tokenization and generation handle more word variations because there is no need to perform stopword replacement or stemming (e.g., for conjugated verbs).

\vspace{-2mm}
\subsection{Structure}
\vspace{-1mm}
\subsubsection{Encoder}
% Confusing. You should follow the same order as the figure:  
% 1. The purpose of the encoder is ...
% 2. The inputs ... are processed by ResNet-50
% 3. ResNet-50 is our base backbone network
% 4. Features are extracted from the 36th layer
The encoder transforms visual information into a latent space feature that is later decoded as a sentence by the decoder. The inputs of the encoder are the target $\textbf{x}_{targ}$, source $\textbf{x}_{src}$ and relation features $\textbf{x}_{rel}$ as illustrated in Fig. \ref{fig:encoder-decoder}. A feature $\textbf{x}_f$ is generated by the encoder.
To do so, the target and source images are both encoded by a convolutional neural network (CNN). In this study, we use ResNet-50~\cite{he2016deep} as the backbone neural network. The encoding process involves extracting the output of the 36$^\text{th}$ layer of ResNet-50, which is followed by a global average pooling (GAP) and a flattening process for dimension reduction.
Feature  $\textbf{x}_f$ is then obtained as the concatenation of the two encoded visual features with the relation feature  $\textbf{x}_{rel}$.
%The purpose of the encoder is converting inputs into features that initialize the decoder. 
%Cropped images of the target and source are processed by ResNet-50~\cite{he2016deep}.
%ResNet-50 is our base backbone network.
%Features are extracted from the 36$^\text{th}$ layer of ResNet-50.
% The encoder network has two sub-encoders and additional input.
% Each sub-encoder is based on ResNet-50 backbone~\cite{he2016deep}.
% and extracts features from $\textbf{x}_{targ}$ or $\textbf{x}_{src}$.
%$\textbf{x}_{targ}$ and $\textbf{x}_{src}$ are processed by ResNet-50 and concatenated to obtain $\textbf{x}_f$
% $\textbf{x}_f$ is obtained by concatenating $\textbf{x}_{rel}$ and the outputs of the two sub-encoders.

%In the original ABN, CNN layers are divided into a base network and a perception branch.
%Our method uses the base network which has the initial $36^{th}$ layers of ResNet-50.
% The reason for using ResNet-50 as CNN was that ResNet-50 outperformed other variations of ResNet in pilot experiments.
%Features of the target $\textbf{x}_{targ}$, the source $\textbf{x}_{src}$ and the full vision image $\textbf{x}_{v}$ are extracted by ResNet-50.
%The target and source features are processed with flatten, concatenated with $\textbf{x}_{rel}$, and used $\textbf{x}_f$ which is the input of the decoder.

\subsubsection{Decoder}
% The decoder is the main part for predicting the next word.
The decoder generates a sequence of latent-space features ${\bf H} = \{{\bf h}_1, {\bf h}_2, \dots, {\bf h}_K\}$, for each step $k$, from the encoded feature $\textbf{x}_f$ by using a multi layer LSTM. These latent-space features allow the linguistic attention and generation branches to predict a sequence of subwords corresponding to the fetching instruction.  For that purpose, each cell of the LSTM, at step $k$, is initialized with the embedding vector of the previous subword predicted ${\bf y}_{k-1}$, as well as a visual feature $\textbf{v}_k$ obtained from the VAB. Feature  $\textbf{v}_k$ is described below with the VAB structure.
%The initial input of the decoder is $\textbf{x}_f$.
%The image feature $\textbf{v}_t$, which is the output of VAB, is input to the decoder cell at each step.
Thereafter, the hidden state of the LSTM propagates as shown in Fig.~\ref{fig:encoder-decoder} and the output ${\bf h}_k$ is generated for each step $k$.

% Inputs of the perception branch are the features concatenated with the LSTM output and the linguistic context masked by the linguistic attention map.

% \begin{table*}[t]
%     \centering
%     \caption{\small Quantitative results on the fetching instruction generation (FIG) task.}
%     \vspace{-1mm}
%     \normalsize
%     \begin{tabular}{l|c c c c c c c}
%         \hline
%          & \multicolumn{7}{c}{Evaluation metric} \\ \cline{2-8}
%         Method & BLEU-1 & BLEU-2 & BLEU-3 & BLEU-4 & ROUGE & METEOR & CIDEr \\ \hline \hline
%         Multi-ABN~\cite{magassouba2019multimodal}(baseline) & 0.550 & 0.389 & 0.274 & 0.191 & 0.412 & 0.209 & 0.453 \\ \hline
%         ABEN (ours) & \textbf{0.605} & \textbf{0.437} & \textbf{0.325} & \textbf{0.240} & \textbf{0.484} & \textbf{0.231} & \textbf{0.723} \\ \hline
%     \end{tabular}
%     \label{tab:result}
%     \vspace{-5mm}
% \end{table*}

\begin{figure}[t]
    \centering
    \includegraphics[clip,width=\linewidth]{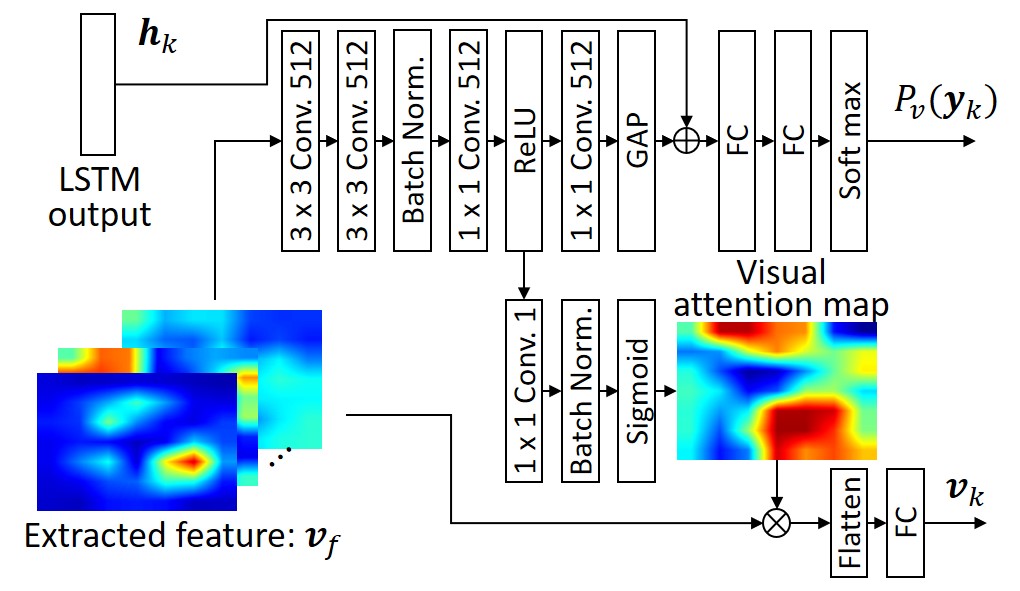}
    \vspace{-8mm}
    \caption{\small  Architecture of the visual attention branch. }
    \label{fig:vab}
    \vspace{-5mm}
\end{figure}

\begin{figure}[t]
    \centering
    \includegraphics[clip,width=\linewidth]{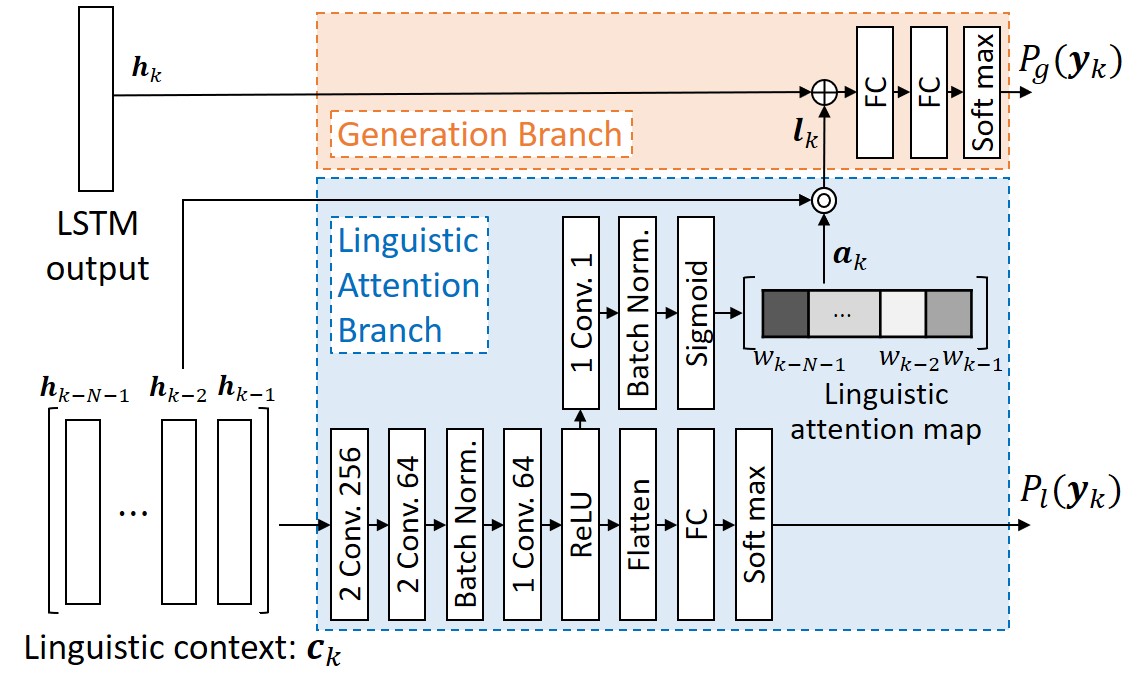}
    \vspace{-7.5mm}
    \caption{\small Architecture of the linguistic and generation branches. $\circledcirc$ is defined as \eqref{equ:w_map}}
    \label{fig:lab}
    \vspace{-2mm}
\end{figure}

\subsubsection{Visual Attention Branch}

Fig.~\ref{fig:vab} shows the structure of the VAB.
The VAB used in this study is based on the method proposed in \cite{magassouba2019multimodal}.  From the VAB, informative regions of features extracted from the image $\textbf{x}_v$ are emphasized to predict the subword ${\bf y}_k$.
 Similarly to the encoder, the input $\textbf{x}_v$ is processed by the 36$^\text{th}$ layer of ResNet-50 and generates feature maps $\textbf{v}_f$.
These feature maps have a dimension $7$~$\times$~$7$~$\times$~$512$ and are input to the VAB after being processed with $2\times2$ average pooling.
The visual feature maps are encoded through four convolutional layers before being processed by a GAP. The likelihood  $P_v({\bf y}_k)$  of the current subword ${\bf y}_k$ is then predicted. In parallel, a visual attention map is created by an additional convolution and sigmoid normalization of the third convolutional layer of the visual attention branch.  This attention map focuses selectively on certain parts of an image related to the predicted sequence. The VAB outputs visual features $\textbf{v}_k$ that are weighted by the attention mask. A cross-entropy loss $L_v$ is minimized by the VAB.

%The visual feature maps weighted by a visual attention mask are used as part of the input for each step of the decoder.

\subsubsection{Linguistic Attention Branch}
Fig.~\ref{fig:lab} shows the network structure of the LAB. The LAB takes, as input, the last $N$ outputs of the LSTM as a linguistic context ${\bf c}_k$.  The parameter $N$ is fixed and is described in the experimental section. 
We define a linguistic context $\textbf{c}_k$ as follows:
\vspace{-0.5mm}
\begin{equation}
   \textbf{c}_k = \{ {\bf h}_{k-N-1}, {\bf h}_{k-N}, \cdots, {\bf h}_{k-1} \},
\vspace{-0.5mm}
\end{equation}
where $\textbf{h}_k$ is the LSTM output at step $k$.
The linguistic context ${\bf c}_k$ has dimension $N\times d$, where $d$ is the dimension of the LSTM hidden state.  Thus, the LAB aims to produce an attention map of dimension 1$\times N$ that weights each component of ${\bf c}_k$.
To this end, ${\bf c}_k$ is processed by three one-dimensional convolutional layers enhanced by batch normalization (BN) and ReLU. Thereafter, the subword ${\bf y}_k$ is predicted from the following fully connected (FC) and softmax layer. In parallel the attention map ${\bf a}_k$ is obtained by connecting the second convolutional layer to a convolutional layer with size 1$\times$1, followed by BN and Sigmoid functions. The attention map ${\bf a}_k$ has dimension 1$\times$N and can be expressed as:
\vspace{-0.5mm}
\begin{equation}
   \textbf{a}_k = \{ w_{k-N-1}, w_{k-N}, \cdots, w_{k-1} \},
\vspace{-0.5mm}
\end{equation}
where each parameter ${w}_k$ is the weight of the corresponding hidden state ${\bf h}_k$. The output
${\bf l}_k$ of the LAB is the weighted linguistic context given by:
\vspace{-0.5mm}
\begin{equation}
   \textbf{l}_k = \{ {\bf o}_{k-N-1}, {\bf o}_{k-N}, \cdots, {\bf o}_{k-1} \},
\vspace{-0.5mm}
\end{equation}
where ${\bf o}_k$ can be expressed as:
\vspace{-0.5mm}
\begin{equation}\label{equ:w_map}
\begin{aligned}
    {\bf o}_{k} &= (1 +  w_{k}) {\bf h}_{k}.
    \end{aligned}
\vspace{-0.5mm}
\end{equation}
Similarly to the VAB, a cross-entropy loss $L_l$ is minimized  based on the likelihood $P_l({\bf y}_k)$ of the predicted subword. 
%$n$ is the number of words in which the method handles previous words.
%The LAB attends a subword within a set of $(k-n-1)$ to $(k-1)$-th subwords when generating $k$-th subword.
%The linguistic attention map ${\bf a}_{k}$ is a 1-dimension vector that emphasizes the current sequence of words that allow predicting correctly the next word.

% To utilize grammatical/syntactical information that requires long sequence,
% To handle the currently generated sequence of words to predict the next word, 

%$\textbf{c}_k$ is expected to handle grammatical/syntactical information and used for the input of the LAB.

% $\textbf{c}_k$ has the dimension $d$ as $\textbf{c}_k = \{\textbf{c}_k^1, \textbf{c}_k^2, \cdots, \textbf{c}_k^d\}$.
% Similarly, $\textbf{l}_k = \{\textbf{l}_k^1, \textbf{l}_k^2, \cdots, \textbf{l}_k^d\}$.
% When $j$ is the index of $d$, the part of the output ${\bf l}_{k}^j$ is obtained from following \eqref{equ:w_map} that is represented as `$ \circledcirc $' in the figure:
% % Weighted feature maps ${\bf l}_{k}$ is obtained through: 
% % To do 
% % this equation is incorrect
% % It should be process each dimension

% After computing with every $j$, $\textbf{l}_k$ is obtained.

% \begin{equation}
% \begin{aligned}
%     {\bf l}_{k}= & \{(1+ w_{k-n-1}) \odot {\bf h}_{k-n-1}, \\
%     & (1+ w_{k-(n-1)-1}) \odot {\bf h}_{k-(n-1)-1}, \\
%     & \cdots, (1+ w_{k-1}) \odot {\bf h}_{k-1}\},
%     \end{aligned}
% \end{equation}

\subsubsection{Generation branch}
Fig.~\ref{fig:lab} shows the structure of the generation branch, which builds the sequence of subwords that compose the fetching instruction. The inputs $\textbf{h}_k$ and $\textbf{l}_k$ are concatenated and processed by FC layers, from which the likelihood of the next subword $p_g(\textbf{y}_k)$ is predicted. A cross-entropy loss $L_g$ is minimized in the generation branch.
%The output of the generation branch is the likelihood $p_g(\textbf{y}_k)$.
%Simple multilayer perceptron (MLP) is used for the generation branch.

\subsubsection{Loss functions}
The global loss function $L_{\text{ABEN}}$ of the network is given by:
\vspace{-0.5mm}
\begin{equation}
    L_{\text{ABEN}} = L_v + L_l + L_g,
\vspace{-0.5mm}
\end{equation}
where $L_v$, $L_l$, and $L_g$ denote cross-entropy losses based on the VAB, LAB, and generation branch, respectively.
Using $L$ as a generic notation for $L_{v}$, $L_{l}$ and $ L_{g}$, the cross-entropy loss is expressed as follows:
\vspace{-0.5mm}
\begin{align} \label{equ:J}
    L &= -\sum_n \sum_{m} y^{*}_{nm} \log p(y_{nm}),
\vspace{-0.5mm}
\end{align}
where $y^{*}_{nm}$ denotes the label given to the $m$-th dimension of the $n$-th sample, and $y_{nm}$ denotes its prediction.  It should be emphasized that the same labels are used for $P_l({\bf y}_k)$ in the LAB and for $P_g({\bf y}_k)$ in the generation branch.

\begin{table}[t]
    \centering
    % \vspace{-6mm}
    \caption{\small Parameter settings of the ABEN}
    \vspace{-2mm}
    \small
    \begin{tabular}{|c|l|}
        \hline
         Opt. & Adam ( Learning rate $=1.0 \times 10^{-4},$ \\
         method & $\beta_1=0.7, \beta_2=0.99999$ ) \\ \hline
        Backbone CNN &  ResNet-50 \\ \hline
        % Embedded vector size &  768 \\ \hline
        LSTM & 3 layers, 768-dimensional cell \\ \hline
        $N$ & 10 \\ \hline
        Generation Branch & FC: 768, 768 \\ \hline
        % MLP num. nodes &  [768, 768] \\ \hline
        % input N size &  downscaled to $(224 \times 224 \times 3)$ \\ \hline
        Batch size & 32 \\ \hline
    \end{tabular}
    \label{tab:parameters}
    \vspace{-6mm}
\end{table}

\begin{table*}[t]
    \centering
    % \color{blue}
    \caption{\small  Quantitative results of FIG. The results are the average over 5 trials. For readability, the metrics are multiplied with 100. ``ABEN w/o BBSE'' uses simple skip-gram instead of BERT-based subword embeddings. ``ABEN (SS)'' and ``ABEN (TF)'' use scheduled sampling and teacher forcing, respectively.}
    \vspace{-3mm}
    \normalsize
    \begin{tabular}{l|c c c c c c c}
        \hline
         & \multicolumn{7}{c}{Evaluation metric} \\ \cline{2-8}
        Method & BLEU-1 & BLEU-2 & BLEU-3 & BLEU-4 & ROUGE & METEOR & CIDEr \\ \hline \hline
        
        VSE~\cite{vinyals2015show} & 43.9$\pm$1.5 & 29.7$\pm$1.3 & 19.0$\pm$1.7 & 11.5$\pm$1.8 & 35.7$\pm$1.2 & 14.3$\pm$0.7 & 21.3$\pm$4.2 \\ \hline

        Multi-ABN~\cite{magassouba2019multimodal}  & 49.1$\pm$0.9  & 35.4$\pm$1.8  & 24.0$\pm$2.3  & 16.0$\pm$2.4  & 37.8$\pm$1.4  & 19.9$\pm$1.1  & 27.5$\pm$5.0  \\ \hline

        ABEN w/o BBSE & 58.2$\pm$1.4  & 38.5$\pm$1.7 & 23.7$\pm$2.6  & 13.9$\pm$2.1  & 42.8$\pm$1.2  & 17.9$\pm$0.6  & 38.0$\pm$2.7   \\ \hline

        ABEN (SS) & {\bf 61.8$\pm$2.8  }& {\bf 46.6$\pm$3.3 }& {\bf 34.0$\pm$3.1} & 24.7$\pm$2.9  & 47.7$\pm$1.4 & 22.0$\pm$1.5 & 54.5$\pm$6.8  \\   \hline
        
        ABEN (TF) & 60.2$\pm$1.9  & 45.1$\pm$1.8  & 33.5$\pm$2.2  & {\bf 24.9$\pm$2.4} & {\bf 48.2$\pm$1.3}& {\bf 22.8$\pm$1.7} & {\bf 57.6$\pm$4.6} \\ \hline
    \end{tabular}
    \label{tab:result}
    \vspace{-3mm}
\end{table*}

\section{Experiments
\label{exp}
}
\vspace{-1mm}
\subsection{Dataset}
\vspace{-1mm}

%The left-hand figure of Fig.~\ref{fig:scene} shows the simulated home environment used in data collection as described in Section \ref{sec:problem}. 
The dataset was collected in simulated home environments as described in Section \ref{sec:problem}. 
The robot patrolled the environment automatically and collected images of designated areas. The environment was procedurally generated with everyday objects and furniture.
%In Fig.~\ref{fig:scene}, typical example of the collected image is shown in the right-hand figure.
%In the image, the target and source are a tumbler and a steel wagon, respectively. 
Each image collected was labeled automatically with the bounding boxes of the sources and targets extracted from the simulator.
These images were then annotated by three different labelers due to the limited size of the dataset. 
% \textcolor{blue}{Based on the standard number of annotators in the field of robotics, these images were annotated by three labelers who were from multiple countries.}
% \textcolor{blue}{In future work we plan to release a bigger dataset with more annotators.}
Each of them was instructed to provide a fetching instruction for each target. 
It should be noted that each image may contain multiple candidate targets and sources. 
Overall, we collected a dataset of 2,865 image--sentence pairs from 308 unique images and 1,099 unique target candidates.  The dataset is available at this
URL\footnote{\protect\href{https://keio.box.com/s/cbup2rttf1gkf487sgad34fqn01wa5r0}{https://keio.box.com/s/cbup2rttf1gkf487sgad34fqn01wa5r0}}. 
% URL\footnote{\protect\url{https://keio.box.com/s/cbup2rttf1gkf487sgad34fqn01wa5r0}}. 

Standard linguistic pre-processing was performed on the instructions. The characters were converted to lowercase, and sentence periods were removed. Stopword replacement and stemming were not performed because subword tokenization and generation were able to handle word variations.

The dataset was split into 80\%, 10\%, and 10\% parts for the training, validation, and test sets, respectively.
After removing invalid samples, we could obtain 2,295 training samples, 264 validation samples, and 306 test samples. 
Because there was no overlap between the training, validation, and test sets, the test set was considered to be unseen.

\vspace{-0.5mm}
\subsection{Parameter settings}
\vspace{-1mm}
The parameter settings of the ABEN are shown in Table~\ref{tab:parameters}.
We used Adam as the optimizer,  with a learning rate of $1 \times 10^{-4}$. The dimension of the BERT-based embedding vector was 768.
We used a three-layer LSTM in the decoder (see  Fig.~\ref{fig:encoder-decoder}) where each cell had a dimension of 768.
The parameter $N$ which characterizes the size of the linguistic context ${\bf c}_k$ was set to 10. More specifically, we considered the 10 previous output of the LSTM to infer the linguistic attention map ${\bf a}_k$. As a result, in the early steps ($k <10$), the linguistic context ${\bf c}_k$ was initialized with the output of the encoder ${\bf x}_f$ for all hidden states that were not available.
% As described in Section~\ref{method}, the linguistic attention of a certain subword is considered only for its former $n$ subwords.
% $n$ was set to 10, therefore $\textbf{c}_j$ was $(10 \times 768)$.
The generation branch had two FC layers, each of which had 768 nodes.
% Every input images (target/source/full images) were downscaled to fit in $(224 \times 224 \times 3)$ before being processed in the decoder and the visual attention branch (VAB).
% Feature $\textbf{x}_{rel}$ was standardized in order to exhibit a zero mean and unit variance for each of them. 
Each dimension of $\textbf{x}_{rel}$ was standardized so that its mean and standard deviation became 0 and 1, respectively.
The visual inputs $\textbf{x}_{targ}$, $\textbf{x}_{src}$ and $\textbf{x}_v$ were resized as $224\times224\times3$ images before being input to ResNet-50.

We trained the ABEN with the aforementioned dataset. The training was conducted on a machine equipped with a Tesla V100 with 32 GB of GPU memory, 768 GB RAM and an Intel Xeon 2.10 GHz processor. The ABEN was trained for 100 epochs, which was sufficient for loss convergence in pilot experiments.

\begin{figure}[t]
    \centering
    \includegraphics[clip,width=70mm]{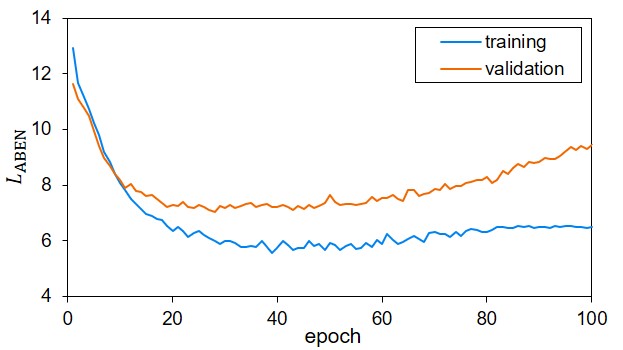}
    \vspace{-3mm}
    \caption{\small  Training and validation losses for 100 epochs.}
    \label{fig:loss}
    \vspace{-3mm}
\end{figure}

\begin{figure*}
    \centering
    \includegraphics[clip,width=\linewidth]{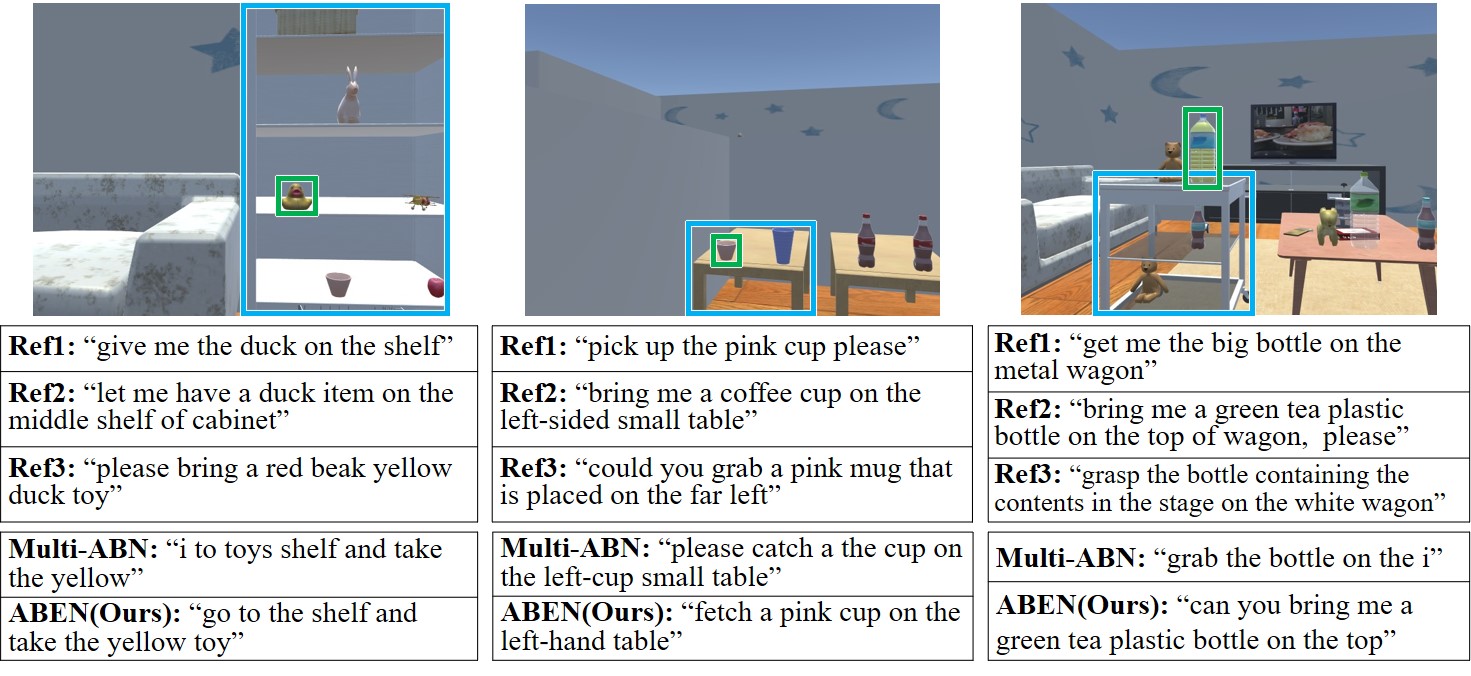}
    \vspace{-8mm}
    \caption{\small Three typical samples of qualitative results. Top figures show input images with bounding boxes of targets and sources. Middle row tables show three reference sentences annotated by labelers. Bottom tables show sentences generated by the Multi-ABN and ABEN.}
    \label{fig:result}
    \vspace{-4mm}
\end{figure*}

\vspace{-1mm}
\subsection{Quantitative results}
\vspace{-1mm}
Table~\ref{tab:result} shows the quantitative results, where standard metrics scores, used in image captioning, are reported.
We conducted five experimental runs for each method. The table shows the mean and standard deviation for each metric.
The BLEU-N column shows the standard BLEU score based on N-grams, where N $\in \{1, 2, 3, 4\}$. 
The CIDEr score was averaged over the same N-grams as BLEU.
Additionally, we used ROUGE-L~\cite{lin2004automatic} which is based on the longest common subsequence. ROUGE-L did not use N-grams.
METEOR was computed from unigrams only, but endows a paraphrase dictionary.

We compared the ABEN with two baseline methods: visual semantic embedding (VSE)~\cite{vinyals2015show}
and Multi-ABN~\cite{magassouba2019multimodal}.
%The Multi-ABN has previously been reported to outperform VSE and speaker \cite{yu2017joint} models.
% Therefore, a comparison was performed only with the Multi-ABN.
%The methods were evaluated through the test set performance when the best validation performance on METEOR was obtained. Hence, these results reflect more objectively the performance on unseen data.
 Based on the standard method for model selection in deep neural network (DNN), we selected the best model as the one that maximized the METEOR score of the validation set. This is because METEOR has a paraphrase dictionary, which is more suitable for handling natural language.
Fig.~\ref{fig:loss} depicts the training and validation loss of a typical run.

The results show that the ABEN outperformed the Multi-ABN and VSE for all four metrics. In particular, the CIDEr score was drastically improved by  30.1 points relative to the Multi-ABN and by 36.3 points relative to, VSE on average. 
 Additionally, the t-test showed that the difference from VSE was statistically significant for all the metrics ($p<0.001$).
The difference from the Multi-ABN was also statistically significant ($p<0.05$). Therefore, the ABEN significantly outperformed these baseline methods for the FIG task. % the p-value for the METEOR was $p=0.013$ and less than $p<0.05$ for all the metrics. Therefore, there was statistically significant difference.
%These results suggests the effectiveness of the proposed BERT-based subword-level embedding and attention model.
% the t-test showed that the results were statistically significant from the Multi-ABN ($p<0.05$) and from VSE ($p<0.001$), for all the metrics.

% 
% In this comparison results between the VSE and ours, the differences were statistically significant ($p<0.001$) for all the metrics. Therefore, the ABEN significantly outperforms the VSE. 
% 

We conducted an ablation study on word embedding. In the ablation, we compared simple skip-gram and BERT-based subword embedding.
In the table, ``ABEN w/o BBSE'' uses simple skip-gram instead of BERT-based subword embedding.
The BERT-based subword embedding has better performance than skip-gram. 
The t-test showed that the results were statistically significant ($p<0.001$) for all the metrics except BLEU-1.

Additionally, we tested two approaches in training: teacher forcing (TF) and scheduled sampling (SS)\cite{bengio2015scheduled}. We adopted the standard SS setup with a linear decay $\epsilon = ({\rm max\_epoch} - {\rm epoch}) / {\rm max\_epoch}$, where $\epsilon$ is the probability of using the label for training.

In the table, the results of these approaches are shown as "ABEN (SS)" and "ABEN (TF)". The t-test showed that the p-values for all the metrics are $p>0.1$. Therefore, there was no statistically significant difference between teacher forcing and scheduled sampling. This indicates that TF did not cause the performance to deteriorate significantly in this task.

\begin{figure}
    \centering
    \includegraphics[clip,width=\linewidth]{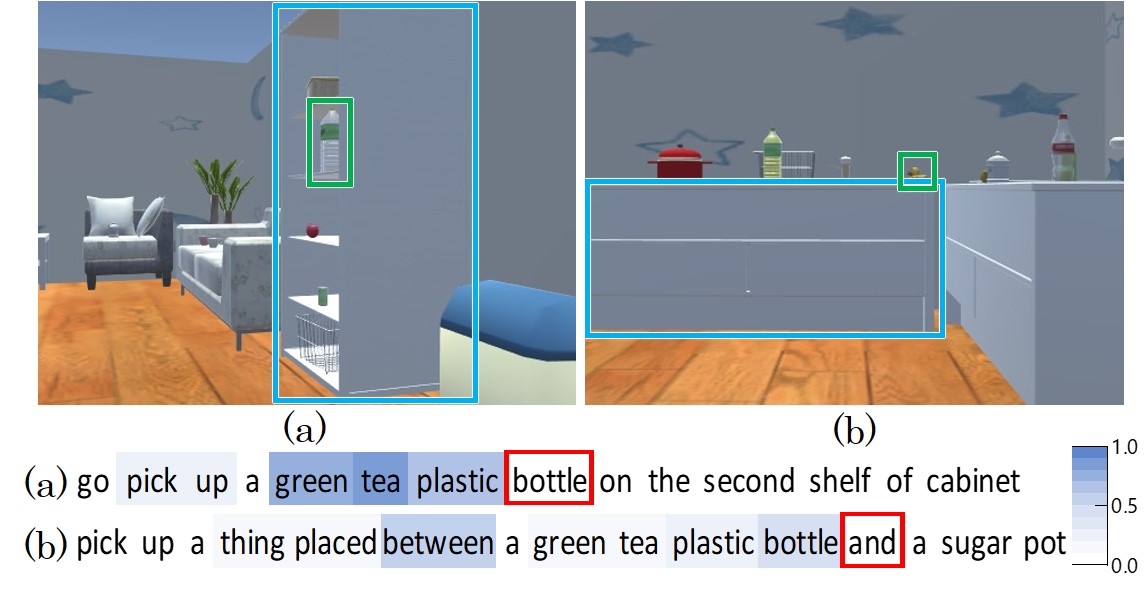}
    \vspace{-6mm}
    \caption{\small  Two typical qualitative results for linguistic attention. The blue and green boxes represent the source and the target. The predicted words are shown in the red boxes, and the attention values are overlaid in blue. In the left case (a), `` grean'', ``tea'' and ``plastic'' were attended for predicting ``bottle''. In the right case (b), ``between'' and ``bottle'' were attended for predicting ``and''. }
    \label{fig:linguistic_attention_map}
    \vspace{-4mm}
\end{figure}

\subsection{Qualitative results}
\vspace{-1mm}
For more insight into the performance of the ABEN, we analyzed the generated sentences qualitatively, as shown in Fig.~\ref{fig:result}.
The top panels of the figures show the input scenes, and the middle and bottom tables show the reference sentences (Ref1, Ref2, and Ref3) and the sentences generated by the ABEN and the Multi-ABN. In the input image, the targets and sources are highlighted by green and blue boxes, respectively.

The left-hand sample in Fig~\ref{fig:result} shows a sentence that was generated successfully, semantically and syntactically, by the ABEN, in contrast to that generated by the Multi-ABN.
Indeed, the target can be uniquely identified from the sentence ``go to the shelf and take the yellow toy'', which is a valid fetching instruction.
Conversely, the sentence generated by the Multi-ABN, refers somehow to the source (`shelf') and the target (`toys', `yellow') but is incorrect syntactically. Such a sentence would require additional review by a human expert in the targeted use case of generating datasets of image--sentence pairs.
Furthermore, the sentence generated by the ABEN is totally different from the reference sentences; this suggests that the many-to-many mapping between language and the environment was captured successfully.

Similarly, the second sample in the middle column illustrates the successful generation of a referring expression, which was used to disambiguate the source.
The ABEN generated the sentence ``fetch a pink cup on the left-hand table'' to refer to the target. In particular, the source was described correctly (``on the left-hand table'') even though the scene contains another similar source. Conversely, the baseline method generated ``please catch a the cup on the left-cup small table'', which included erroneous syntax about the source (``left-cup'' instead of ``left-hand''). Additionally, over-generation appeared, as the phrase ``a the''.
% The sentence generated by Multi-ABN was not so worse a result but included bad word, ``left-cup''.

% In contrast, the ABEN generated the sentence correctly including a referring expression ``a pink cup on the left-hand table''.
% The sentences of Ref2 and Ref3 include the spatial referring expressions ``a coffee cup on the left-sided small table'' and ``a pink mug that is placed on the far left''.
% Both referring expressions describe the left-hand table, but their word orders were completely different.

% Despite, the ABEN can deal with the referring expression ``on the left-hand table''.

% In the right-hand column of the figure, The baseline method generated erroneous expression ``grab the bottle on the i''.
% The sentence included the non-unique target name ``the bottle'' and was missing referring expressions due to generation failure.
% In this image, there were several bottles, therefore, the sentence should included referring expressions to determine the target uniquely.
% The generated sentence of ABEN was more natural with detail information such as ``green tea plastic bottle'' and ``on the top''.
% The target can be determined from this instruction because the target object was the top of these bottles.
% However, in the sentence, the landmark object name might be missing, for instance ``on the top of the white wagon''.
% We assumed that the cause of the missing words was under-generation of LSTM.

The right-hand sample illustrates ambiguity about the target. In this scene, there were three bottles. Therefore, the sentence should include referring expressions to determine the target uniquely. The sentence generated by the ABEN was able to disambiguate the target, which was referred to as ``green tea plastic bottle''. However, the source description was incomplete. Indeed, a more exhaustive source description such as ``on the top of the white wagon'', could be expected. Nonetheless, this fetching instruction remains understandable to human experts. Conversely, the baseline method generated a sentence that was syntactically incorrect but also ambiguous. The target was simply referred to as ``bottle'', from which the target cannot be identified. 

% was more natural with detail information such as  and ``on the top''. column of the figure shows the failure results of the ABEN.
% The target was the apple on the shelf.
% The baseline method generated an erroneous sentence ``i want a apple an apple an on placed in cola''.
% There were over-generation ``a apple an apple an on'' and erroneous a referring expression ``placed in cola''.
% The referring expression should be ``an apple placed in the shelf''.
% The generated sentence of the ABEN was ``take the fruit on the table''.
% The result was also mistaken due to incorrect source name ``table''.
% In the scene image, the source might be difficult to be classified because the lower part of the shelf was hidden by the sofa.
% Additionally, the shelf was an unnatural white plane and has extremely few surface features.
% % The angle of the view was also challenging.
% For the above reasons, we considered that the source name did not appear in the generated sentence.
% In future work, we will perform in a more photo-realistic or actual environment.

% Multi-ABN's result shown in the bottom row was basically correct.
% However, there are two stuffed dog in the input image.
% These cannot be distinguished by the instruction.
% On the other hand, the generated sentence of ABEN includes referring expression of ``on the kitchen cabinet of the blue'' to distinguish them.

Additionally, we analyzed the relationship between the subwords in the sentence generation process through the linguistic attention maps shown in Fig.~\ref{fig:linguistic_attention_map}.
The lower part illustrate the salient subwords that were used to predict the subword marked with a red frame.
In Fig.~\ref{fig:linguistic_attention_map}(a), to predict ``bottle'', the most salient subwords were ``green'', ``tea'', and ``plastic''.
%It is also shown that "pick" and "up" had a slight relationship with "bottle".
In Fig.~\ref{fig:linguistic_attention_map}(b), to predict the subword ``and'' in the sentence, the most salient words were "bottle" but also ``between''.
These results indicate that the ABEN handles subword relationships in a representation that is understandable to humans.

Overall, these results emphasize that the ABEN generates more natural sentences than the baseline method, through the contribution of our proposed LAB architecture and the subword generation strategy. 
%From the above qualitative results, the ABEN can be more reasonable than the Multi-ABN.

% 
% \subsection{Discussions}

% \subsection{About dataset bias}
% Labeling our dataset with more and various annotators,  is preferable to avoid any linguistic bias. Nonetheless, our dataset was labeled by annotators from two different countries and they have different backgrounds to avoid such bias.

% In NLP, it is true that datasets are often annotated by more than ten annotators. However, in robotics, datasets are annotated by one to three people in most studies (e.g. \cite{nyga2018grounding,mees2020learning,antunes2019bidirectional}).

% Nonetheless in our future work, we are planning to add more annotators to build a large scale dataset. 
% 
% \vspace{-2mm}
\section{Conclusions}
% \vspace{-1mm}
Most data-driven approaches for multimodal language understanding require large-scale datasets.  
However, building such a dataset is time-consuming and costly.
In this study, we proposed the ABEN, which generates fetching instructions from images. Target use cases include generating and augmenting datasets of image--sentence pairs.

The following contributions of this study can be emphasized:
\begin{itemize}
 \item The ABEN extends the  Multi- ABN by introducing a linguistic branch and a generation branch,  to model the relationship between subwords. 
 \item  The ABEN combines attention branches and BERT-based subword embedding for sentence generation. 
\end{itemize}

Future studies will investigate the application of the ABEN to real-world settings.

% \appendix
% \input{sectionA}

\vspace{-1mm}
\section*{ACKNOWLEDGMENT}
\vspace{-0.5mm}
This work was partially supported by JSPS KAKENHI Grant Number 20H04269, JST CREST, SCOPE, and NEDO.
\vspace{-0.5mm}
\bibliographystyle{IEEEtran}
\bibliography{reference}

\end{document}